\documentclass{article}

\usepackage{PRIMEarxiv}

\usepackage[utf8]{inputenc} % allow utf-8 input
\usepackage[T1]{fontenc}    % use 8-bit T1 fonts
\usepackage{hyperref}       % hyperlinks
\usepackage{url}            % simple URL typesetting
\usepackage{booktabs}       % professional-quality tables
\usepackage{amsfonts}       % blackboard math symbols
\usepackage{nicefrac}       % compact symbols for 1/2, etc.
\usepackage{microtype}      % microtypography
\usepackage{lipsum}
\usepackage{fancyhdr}       % header
\usepackage{graphicx}       % graphics
\graphicspath{{media/}}     % organize your images and other figures under media/ folder

\usepackage{hyperref}
\usepackage{url}
\usepackage{multirow} 
\usepackage{hyperref}
\usepackage{graphicx}
\usepackage{url}

\usepackage{algorithm}
\usepackage{times}
\usepackage{epsfig}
\usepackage{amsmath}
\usepackage{amssymb}
\usepackage{algpseudocode}
\usepackage{amsthm}
\usepackage{amsfonts,bm}

\usepackage[table]{xcolor} % 用于设置行背景颜色
\usepackage{array} % 用于调整列间距
\usepackage{adjustbox} % 用于调整表格大小

\usepackage{booktabs}
\usepackage{bbding}
\usepackage{epigraph}
\usepackage{tcolorbox}
\usepackage{colortbl, xcolor}

\definecolor{grey}{rgb}{0.1,0.1,0.1} 
\usepackage{pifont}

\title{Learning New Concepts, Remembering the Old: Continual Learning for Multimodal Concept Bottleneck Models }
\author{
  Songning Lai \\
  HKUST(GZ) \\
  Deep Interdisciplinary Intelligence Lab\\
  \texttt{songninglai@hkust-gz.edu.cn} \\
   \And
  Mingqian Liao \\
  HKUST(GZ) \\
  Deep Interdisciplinary Intelligence Lab\\
  \texttt{mliao529@connect.hkust-gz.edu.cn} \\
     \And
  Zhangyi Hu \\
  HKUST(GZ) \\
  Deep Interdisciplinary Intelligence Lab\\
  Wuhan University\\
  \texttt{zhangyihu@hkust-gz.edu.cn} \\
       \And
  Jiayu Yang \\
  HKUST(GZ) \\
  Deep Interdisciplinary Intelligence Lab\\
\texttt{jyang729@connect.hkust-gz.edu.cn} \\
       \And
  Wenshuo Chen \\
  HKUST(GZ) \\
  Deep Interdisciplinary Intelligence Lab\\
  Shandong University\\
\texttt{chatoncws@gmail.com} \\
       \And
  Hongru Xiao \\
  Tongji University \\
\texttt{hongru\_xiao@tongji.edu.cn} \\
       \And
  Jianheng Tang \\
  Peking University \\
  Deep Interdisciplinary 
\texttt{tangentheng@gmail.com} \\
       \And
  Haicheng Liao \\
  University of Macau \\
\texttt{yc27979@um.edu.com} \\
  \And
  Yutao Yue\thanks{Correspondence to Yutao Yue \{yutaoyue@hkust-gz.edu.cn\}}. \\
  HKUST(GZ) \\
  Institute of Deep Perception Technology, JITRI\\
  Deep Interdisciplinary Intelligence Lab\\
  \texttt{yutaoyue@hkust-gz.edu.cn}\\
  \thanks{This work was supported by Guangzhou-HKUST(GZ) Joint Funding Program (Grant No.2023A03J0008), Education Bureau of Guangzhou Municipality.}
}

\begin{document}
\maketitle

\begin{abstract}
Concept Bottleneck Models (CBMs) enhance the interpretability of AI systems, particularly by bridging visual input with human-understandable concepts, effectively acting as a form of multimodal interpretability model. However, existing CBMs typically assume static datasets, which fundamentally limits their adaptability to real-world, continuously evolving multimodal data streams. To address this, we define a novel continual learning task for CBMs: simultaneously handling \textbf{concept-incremental and class-incremental learning}. This task requires models to continuously acquire new concepts (often representing cross-modal attributes) and classes while robustly preserving previously learned knowledge. To tackle this challenging problem, we propose \textbf{CON}ceptual \textbf{C}ontinual \textbf{I}ncremental \textbf{L}earning (\textbf{CONCIL}), a novel framework that fundamentally re-imagines concept and decision layer updates as linear regression problems. This reformulation eliminates the need for gradient-based optimization, thereby effectively preventing catastrophic forgetting. Crucially, CONCIL relies solely on recursive matrix operations, rendering it highly computationally efficient and well-suited for real-time and large-scale multimodal data applications. Experimental results compellingly demonstrate that CONCIL achieves “absolute knowledge memory” and significantly surpasses the performance of traditional CBM methods in both concept- and class-incremental settings, thus establishing a new paradigm for continual learning in CBMs, particularly valuable for dynamic multimodal understanding.
\end{abstract}
% \vspace{-12pt}
\section{Introduction}
\label{sec:intro}
% \vspace{-2pt}
%%%%% 

Deep learning has profoundly transformed numerous fields, achieving unprecedented performance across a diverse range of tasks, particularly in multimodal data processing such as computer vision, natural language processing, and audio analysis \cite{lecun_deep_2015, Samek_2021}. However, despite these remarkable advancements in model accuracy and scalability, a critical challenge persists: interpretability. This limitation has spurred the emergence of Explainable Artificial Intelligence (XAI), a research area dedicated to developing models whose decision-making processes are understandable to humans. Within XAI, Concept Bottleneck Models (CBMs) have emerged as a highly promising approach for interpretable multimodal AI. By explicitly embedding human-understandable concepts (e.g., visual attributes like ‘striped,’ ‘round,’ or ‘metallic’) into their architectural structure, CBMs inherently bridge distinct modalities: mapping complex raw input (e.g., an image) to interpretable, often textual or symbolic, concept explanations. This enables users to not only interpret but also intervene in the model’s underlying reasoning, crucial for understanding decisions based on complex multimodal content (e.g., classifying images based on inferred attributes, segmenting video through object concepts, or analyzing audio events via their semantic descriptors) \cite{9552218}. Such cross-modal interpretability is indispensable in high-stakes domains like medical imaging diagnostics (interpreting image features with clinical terms), autonomous vehicle perception (explaining visual scene understanding with abstract concepts), and financial fraud detection from multimodal data (linking disparate data sources to understandable risk factors), where comprehending the “why” behind a model’s decision can be as crucial as the decision itself \cite{Longo_2024}.

\begin{figure*}[t]
\centering
% \vspace{-7pt}
\includegraphics[width=0.8\textwidth]{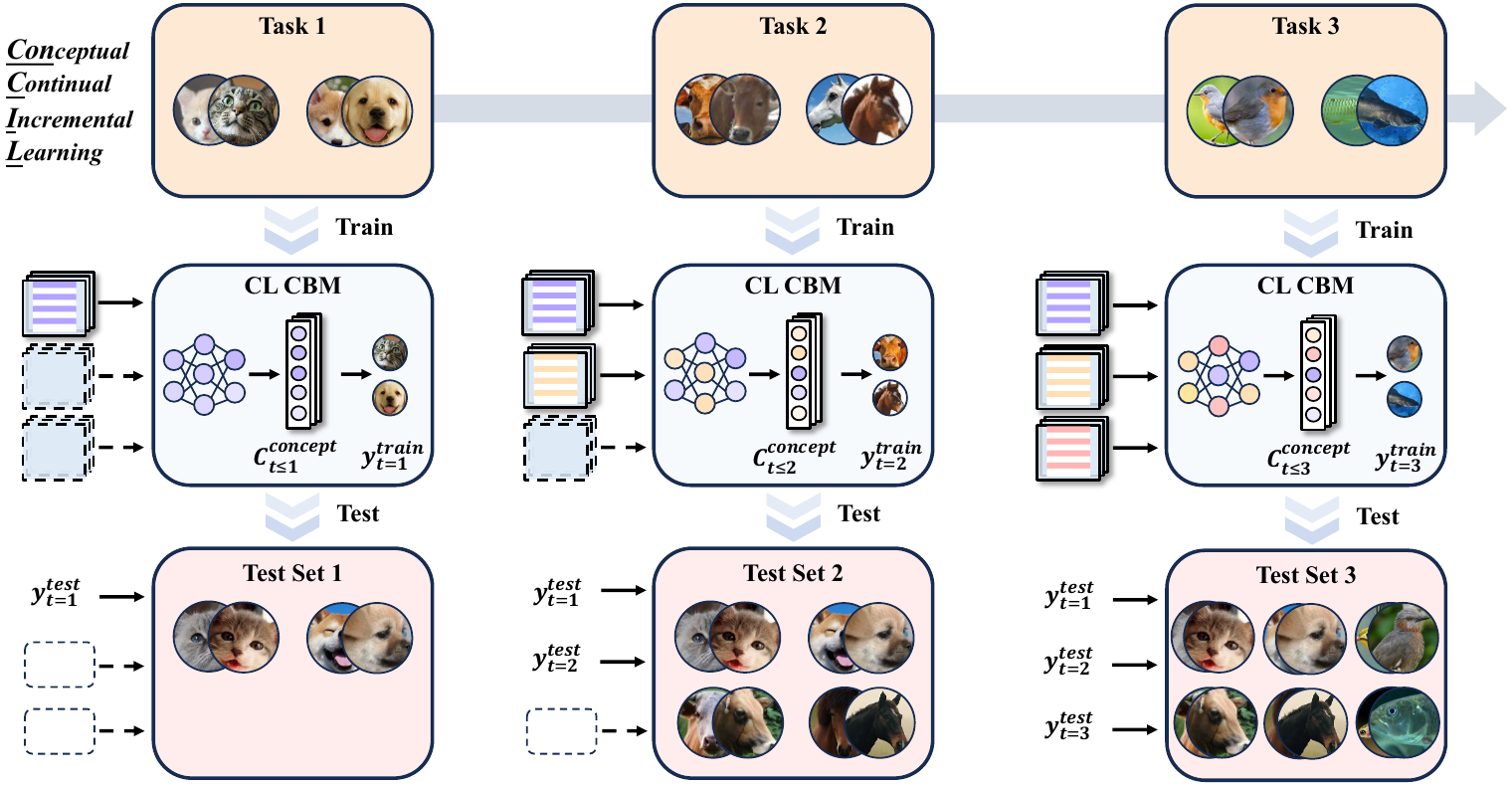}
% \vspace{-10pt}
\caption{Schematic illustration of the Concept-Incremental and Class-Incremental Continual Learning (CICIL) task for CBMs. The figure depicts the sequential nature of the CICIL task, where each task $ t $ introduces new classes and concepts. The training and testing datasets $\mathcal{D}_t^{\text{train}}$ and $\mathcal{D}_t^{\text{test}}$ for each task contain input feature vectors $\mathbf{x}$, concept vectors $\mathbf{c}$, and labels $y$. The cumulative concept set $\mathcal{C}^{\text{concept}}_{\leq t}$ expands over time, incorporating both previously learned and new concepts. }
\label{Intro-CONCIL}
% \vspace{-25pt}
\end{figure*}

Despite the significant strides in CBM research, a substantial gap remains regarding their deployment in real-world environments, where multimodal data is inherently dynamic rather than static. Most existing CBM studies, like much of traditional machine learning, operate under the assumption of static datasets \cite{wu2024continuallearninglargelanguage}, neglecting the fluidity of real-world multimodal streams. In practice, however, multimodal data (e.g., surveillance video feeds combining visual and audio, social media content with image-text pairs, sensor data from IoT devices integrating various modalities) is continuously collected and updated incrementally over time, forming perpetual multimodal data streams characterized by shifting distributions, the emergence of new classes (e.g., new object categories in video, new audio events, novel image-text alignments), and the evolution of underlying concepts. To effectively operate in such scenarios, models must not only learn proficiently from novel information but also robustly retain previously acquired knowledge. This necessity introduces \textbf{continual learning (CL)}, a paradigm that enables models to adapt to new information without suffering from catastrophic forgetting of prior knowledge \cite{wang2024comprehensivesurveycontinuallearning}. Nevertheless, a pervasive challenge in traditional deep learning approaches is that acquiring new information often leads to the severe degradation or complete loss of prior knowledge—a phenomenon known as catastrophic forgetting \cite{MCCLOSKEY1989109, McClelland1995WhyTA, Kirkpatrick_2017}.

To date, no existing research has explored the integration of continual learning within CBM frameworks, particularly in the challenging context of simultaneously handling \textbf{concept-incremental} and \textbf{class-incremental} tasks \cite{van_de_ven_three_2022}. This represents a critical and unaddressed gap, as real-world multimodal data is inherently dynamic, necessitating intelligent systems capable of accommodating both the arrival of new predictive categories (classes) and the evolution of underlying explanatory concepts. The pressing need for such integration can be clearly illustrated through practical examples, particularly in multimodal AI applications. Consider a medical diagnostic system powered by CBMs that analyzes multimodal medical data (e.g., medical images like X-rays or MRI scans, alongside patient reports and audio notes): as new diseases (classes) emerge and additional symptoms or diagnostic markers (concepts, e.g., ‘calcification patterns’ from X-rays, ‘edema characteristics’ from MRI, ‘dyspnea’ from text/audio, bridging visual, textual, and audio modalities) are identified, the system must seamlessly learn from these new developments while fully retaining its diagnostic capabilities for previously learned diseases and symptoms. Similarly, in advanced surveillance systems processing continuous multimodal streams (e.g., video, thermal, audio), new object categories (classes, e.g., ‘new type of drone’) or behavioral concepts (e.g., ‘suspicious loitering pattern,’ ‘distress vocalization’) may emerge over time, demanding continuous adaptation and transparent reasoning across modalities. More importantly, while new data phases introduce new classes, they also bring new concepts alongside a continuation of previously learned concepts. The model must retain its understanding of all observed concepts, even if the classes they were originally associated with do not explicitly recur in subsequent learning phases. This intricate scenario necessitates both \textbf{concept-incremental} learning (where the set of understood concepts cumulatively expands over time) and \textbf{class-incremental} learning (where new classes are introduced per task, but the classifier is typically adapted to classify only the current and previously seen classes, or only the current classes), forming a complex and multifaceted learning task that has, to our knowledge, not been addressed in the current CBM literature.

In this work, we define for the \textbf{first time} a novel \textbf{concept-incremental and class-incremental continual learning task specifically within the CBM framework}, a challenge that precisely mirrors the complexities of real-world multimodal data collection and labeling practices. In this task, while new data at each phase introduces distinct sets of classes for prediction, the underlying concepts that the model must learn are cumulative across all phases—they expand and build upon concepts acquired in previous phases. Crucially, \textbf{new data for each task contains not only new concepts but also instances that leverage previously learned concepts}, which the model must effectively utilize to maintain its accuracy while adapting to evolving knowledge. This multi-phase, concept-incremental and class-incremental learning task is not only more closely aligned with practical multimodal data evolution but also poses significant challenges for existing CBM models, which were not designed to handle this unique form of incremental learning.

Given the complete absence of solutions for this novel and complex task, we propose \textbf{the first continual learning framework for CBMs}, meticulously designed to address the unique challenges of concept-incremental and class-incremental learning. Our approach operates under realistic constraints, particularly relevant for large-scale multimodal AI systems: at each learning phase, only the data pertinent to the current phase and the model weights from the immediate previous phase are available. This constraint faithfully reflects common data privacy regulations and storage limitations encountered in real-world deployments involving continuous multimodal data streams. Traditional continual learning strategies typically involve either computationally expensive retraining on all historical data (which may also violate privacy) or risky fine-tuning solely on new data \cite{lin2024mitigatingalignmenttaxrlhf}, which inevitably leads to catastrophic forgetting \cite{luo2024empiricalstudycatastrophicforgetting}. In stark contrast, our framework leverages \textbf{analytic learning principles} \cite{Gogus2012} to fundamentally reformulate the updates required for the CBM’s concept and decision layers as a series of linear regression problems. This innovative reformulation eliminates the dependence on gradient-based updates, thereby inherently and effectively preventing catastrophic forgetting, ensuring that the model retains prior knowledge while seamlessly adapting to new concepts and classes.

To instantiate this paradigm, we introduce \textbf{CON}ceptual \textbf{C}ontinual \textbf{I}ncremental \textbf{L}earning (\textbf{CONCIL}), a framework that exclusively utilizes lightweight recursive matrix operations for all model updates. This design makes CONCIL exceptionally computationally efficient and inherently scalable, rendering it highly suitable for real-time applications and processing large-scale, continuous multimodal data streams. We provide a rigorous theoretical proof demonstrating that CONCIL achieves “absolute knowledge memory,” implying that the model effectively behaves as though it had been trained on a centralized, comprehensive dataset encompassing all past and current knowledge, thus fully retaining knowledge from previous phases. Through extensive empirical validation, we compellingly demonstrate that \textbf{CONCIL} significantly outperforms traditional CBM methods in both concept-incremental and class-incremental settings, showcasing its superior adaptability and knowledge retention capabilities.

The key contributions of this paper are summarized as follows:
\begin{itemize}
% % \vspace{-5pt}
\item We define and rigorously formulate the \textbf{concept-incremental and class-incremental continual learning task within the CBM framework} for the \textbf{first time}, identifying a novel and critical challenge that accurately reflects real-world multimodal data evolution.
\item We propose the \textbf{first continual learning framework for CBMs} that fundamentally eliminates the need for gradient-based updates by reformulating layer updates as analytic linear regression problems, thereby inherently and effectively preventing catastrophic forgetting and ensuring robust retention of prior knowledge.
\item We introduce \textbf{CONCIL}, an innovative framework based on highly efficient recursive matrix operations, making it computationally lightweight, scalable, and practical for real-time and large-scale multimodal continual learning applications.
\item We provide comprehensive theoretical guarantees, including a proof of “absolute knowledge memory,” and extensive empirical validation demonstrating CONCIL’s superior effectiveness in the context of continual learning tasks within CBMs compared to traditional approaches.
% \vspace{-10pt}
\end{itemize}
%%% \vspace{-5pt}
\section{Related Work}
\label{sec:rel}
%%% \vspace{-5pt}

\noindent\textbf{Concept Bottleneck Models (CBMs)} are a prominent class of Explainable Artificial Intelligence (XAI) techniques designed to enhance model interpretability by integrating human-understandable concepts as intermediate representations within neural network architectures. These models often serve as multimodal interpretability bridges, mapping raw input from one modality (e.g., images, video frames, audio clips) to high-level, interpretable concepts (e.g., visual attributes, object categories, action verbs), which can often be expressed textually or symbolically. The foundational CBMs \cite{koh2020concept} significantly improved transparency by embedding concept-based layers. Subsequent advancements have further diversified CBM functionalities: Interactive CBMs \cite{chauhan2023interactive} aim to improve predictive accuracy through interactive concept learning, while Post-hoc CBMs (PCBMs) \cite{yuksekgonul2022post} and Label-free CBMs \cite{oikarinen2023label} enable interpretability for pre-trained models and facilitate unsupervised learning without explicit concept annotations, respectively. While these approaches collectively contribute to the interpretability landscape, a critical limitation shared by virtually all existing CBMs is their predominant assumption of static datasets. This assumption severely constrains their applicability in real-world scenarios where multimodal data arrives in continuous, dynamic streams, characterized by evolving concepts and classes. To address this fundamental limitation, our work introduces the first continual learning paradigm specifically tailored for CBMs, designed to inherently handle both concept-incremental and class-incremental learning scenarios for dynamic multimodal environments. This novel paradigm marks a significant conceptual departure from prior static data assumptions, enabling CBMs to effectively adapt to perpetually evolving multimodal data environments.

\noindent\textbf{Class-Incremental Learning (CIL)} has garnered substantial attention as a crucial area within continual learning, specifically focusing on mitigating catastrophic forgetting in models trained on sequentially arriving data \cite{CIL_survey, LwF, FASTICARL}. CIL methods are broadly categorized into replay-based and exemplar-free approaches. Replay-based methods, such as iCaRL \cite{rebuffi2017icarl} and ER \cite{ER}, aim to preserve prior knowledge by storing and replaying a subset of previously encountered samples during subsequent incremental training. However, this strategy often faces practical hurdles due to privacy concerns and substantial storage requirements for historical data, especially when dealing with large volumes of multimodal exemplars. In contrast, exemplar-free approaches \cite{LwF, MAS, DT2W} endeavor to safeguard prior knowledge through regularization or knowledge distillation techniques, without retaining past samples. While these methods circumvent storage issues, they often exhibit comparatively less effective performance than their replay-based counterparts. Critically, both categories predominantly rely on gradient-based updates, which, despite various mitigation strategies, can still inadvertently lead to the degradation of previously learned information when new data is introduced. This issue is compounded in multimodal learning where complex feature representations across modalities are highly intertwined. Distinctly, our proposed framework for CBMs diverges from this reliance, employing a novel gradient-free, analytic learning approach specifically designed to prevent catastrophic forgetting while ensuring exceptional computational efficiency, making it highly suitable for evolving multimodal tasks.

\noindent\textbf{Analytic Learning} provides a powerful alternative to traditional gradient-based optimization by directly computing model parameters through closed-form mathematical solutions, such as least-squares regression \cite{AL_1, AL_2}. Although analytic learning holds significant promise for continual learning by inherently reducing computational costs and mitigating catastrophic forgetting, its application to scenarios involving incrementally evolving concepts or classes, particularly within multimodal contexts, has not been widely explored. Notable examples include radial basis function networks \cite{AL_3}, which leverage least-squares estimation following kernel transformations, and multilayer analytic learning \cite{AL_4}, which efficiently segments nonlinear learning tasks for single-epoch training. The inherently gradient-free nature of analytic learning offers a robust theoretical foundation for developing continual learning solutions, particularly for tasks demanding the absolute retention of knowledge across incrementally added concepts and classes. This characteristic is especially advantageous for multimodal data streams where gradient propagation across complex, inter-dependent modalities can be challenging and computationally expensive. Building on these principles, our framework pioneers the application of recursive matrix operations to enable memory-preserving model updates in CBMs, rendering it uniquely suitable for real-time applications and large-scale, dynamic multimodal data processing.

While significant advancements have been made in enhancing interpretability in CBMs (including their emerging role in multimodal understanding), developing strategies to mitigate forgetting in CIL, and exploring the potential of analytic learning as a non-gradient-based optimization paradigm, no existing work has successfully integrated these disparate research threads to address the highly complex and unique challenge of continual learning within CBM frameworks, particularly for dynamic multimodal data. Our work unequivocally fills this critical void. We uniquely define the task of \textbf{concept-incremental and class-incremental continual learning specifically for CBMs} and present a precisely tailored analytic solution to this multifaceted problem. This groundbreaking approach enables CBMs to adapt robustly to evolving multimodal data streams without any sacrifice of prior knowledge, thereby establishing a novel and crucial benchmark for interpretable models operating in dynamic real-world environments, including multimodal AI applications.
% \vspace{-3pt}
\section{Preliminary}
\label{sec:preli}
% \vspace{-3pt}

% \vspace{-2pt}
\subsection{Concept Bottleneck Model}
% \vspace{-2pt}

We introduce Concept Bottleneck Models (CBMs) using the established notation by Koh et al. \cite{koh2020concept}. In a CBM, a predefined set of high-level, human-understandable concepts, denoted as $\mathcal{C} = \{c^1, \ldots, c^L\}$, serves as an intermediate, interpretable representation within the model architecture. The training dataset is typically given as $\{(\mathbf{x}_i, \mathbf{c}_i, y_i)\}_{i=1}^{n}$, where for each instance $i \in [n]$, $\mathbf{x}_i \in \mathbb{R}^d$ is the input feature vector, $\mathbf{c}_i \in \mathbb{R}^L$ is the corresponding concept vector (where each element $c^k$ represents the $k$-th concept for input $\mathbf{x}_i$), and $y_i \in \mathbb{R}$ is the class label.

CBMs are structured to learn two sequential mappings from this dataset. The first mapping, $g: \mathbb{R}^d \rightarrow \mathbb{R}^L$, known as the concept extractor, projects the input feature vector $\mathbf{x}_i$ into the concept space, yielding a predicted concept vector $\hat{\mathbf{c}} = g(\mathbf{x})$. The second mapping, $f: \mathbb{R}^L \rightarrow \mathbb{R}$, serving as the classifier, then utilizes these predicted concepts to infer the target label, resulting in $\hat{y} = f(\hat{\mathbf{c}}) = f(g(\mathbf{x}))$. The core objective in training CBMs is to ensure that both the predicted concept vector $\hat{\mathbf{c}}$ and the final class prediction $\hat{y}$ accurately approximate their respective ground truth values, $\mathbf{c}$ and $y$. This architectural design allows CBMs to provide transparency in model decision-making by enabling users to interpret intermediate representations in terms of human-understandable concepts, thereby facilitating potential intervention based on concept-level reasoning.

Due to space limitations, related work about CBMs, CIL, and analytic learning are chapters in Appendix \ref{sec:rel}.

% \vspace{-7pt}
\section{Task Definition: Concept-Incremental and Class-Incremental Continual Learning (CICIL)}
\label{sec:taskdefi}
% \vspace{-3pt}

We formally define the Concept-Incremental and Class-Incremental Continual Learning (CICIL) task specifically for CBMs, as schematically illustrated in Figure \ref{Intro-CONCIL}. The learning process unfolds across $T$ sequential tasks, indexed by $t \in \{1, \dots, T\}$. Each task $t$ introduces a set of new classes that are disjoint from previously observed classes, and crucially, it involves both previously learned concepts and new concepts unique to the current task.

For each task $t$, we are provided with a training dataset $\mathcal{D}_t^{\text{train}} = \{(\mathbf{x}_i, \mathbf{c}_i, y_i)\}_{i=1}^{N_t}$ and a corresponding testing dataset $\mathcal{D}_t^{\text{test}} = (\mathbf{x}_j, \mathbf{c}_j, y_j), \\
j=1,\dots,M_t$. Here, $\mathbf{x}_i \in \mathbb{R}^d$ represents the input feature vector, $\mathbf{c}_i \in \mathbb{R}^{L_t}$ is the concept vector relevant for task $t$, and $y_i \in Y_t$ is the class label associated with task $t$. A defining characteristic of this task is that the sets of classes for different tasks are mutually exclusive, i.e., $Y_t \cap Y_{t'} = \emptyset$ for $t' \neq t$.

Furthermore, each task $t$ is associated with a specific set of concepts $\mathcal{C}_t^{\text{concepts}} = \{c^1, \ldots, c^{L_t}\}$. This set is composed of a subset of concepts that have already been learned in prior tasks, as well as novel concepts introduced for the first time in task $t$. Consequently, at the completion of task $t$, the cumulative set of concepts understood by the model is $\mathcal{C}^{\text{concepts}}_{\leq t} := \mathcal{C}^{\text{concepts}}_{\leq t-1}\cup  \mathcal{C}^{\text{concepts}}_{t}$. This formulation mandates that the CBM must not only incrementally acquire the ability to classify new classes introduced in each task but also continuously expand and retain its concept representations over the entire learning sequence. After the completions of $T$ tasks, we get the updated concepts set $\mathcal{C}^{\text{concepts}}_{\leq t}$, where $\#\mathcal{C}^{\text{concepts}}_{\leq t} = \sum_{t=1}^{T}\#\mathcal{C}^{\text{concepts}}_{t}$.

In this continual learning setting, the CBM architecture remains consistent, comprising two main components: (1) a concept extractor $g: \mathbb{R}^d \rightarrow \mathbb{R}^{L_{\leq t}}$ that maps the input feature space to the *accumulated* concept space (which grows with each task), and (2) a classifier $f: \mathbb{R}^{L_{\leq t}} \rightarrow \mathbb{R}^{|Y_{\leq t}|}$ that maps the accumulated concept space to the *cumulative* class label space (which also grows). For a given input $\mathbf{x}$, the predicted concept vector at task $t$ is $\hat{\mathbf{c}} = g(\mathbf{x})$, and the final predicted label is $\hat{y} = f(g(\mathbf{x}))$.

\noindent\textbf{Objective in Concept- and Class-Incremental Learning. }At each sequential task $t$, the primary objective is to learn the updated model parameters $\Theta_t = \{\theta_g, \theta_f\}$ for the concept extractor $g$ and classifier $f$. This learning process leverages both the new dataset $\mathcal{D}_t^{\text{train}}$ and the model parameters from the previous task, $\Theta_{t-1}$. The updated model $\Theta_t$ must rigorously satisfy two critical, often conflicting, properties essential for continual learning:

\noindent\textbf{(i) Stability (Knowledge Retention):} The model must robustly preserve its ability to accurately predict concepts and classes learned in all preceding tasks, from $1$ to $t-1$. This property is paramount for ensuring that accumulated knowledge is fully retained and for mitigating the detrimental effects of catastrophic forgetting.

\noindent\textbf{(ii) Plasticity (Adaptability to New Information):} The model must exhibit sufficient adaptability to effectively learn and integrate new concepts and classes introduced in the current task $t$. This demands that the concept extractor $g$ can recognize and form representations for novel concepts, while the classifier $f$ can accurately distinguish new classes without experiencing negative interference or degradation in performance on previously learned classes.

A crucial constraint in this framework is that the model is limited to accessing only the current task's training data $\mathcal{D}_t^{\text{train}}$ and the parameters from the immediately preceding task $\Theta_{t-1}$. This constraint directly reflects realistic deployment scenarios where access to all historical data is often impractical or prohibited due to significant storage requirements, computational overhead, or strict data privacy regulations.

The simultaneous increment of both concepts and classes in our defined task presents unprecedented architectural and learning challenges. Specifically, the output dimension of the concept extractor $g$ (corresponding to the growing concept space) and both the input and output dimensions of the classifier $f$ (corresponding to the growing concept space and growing class space, respectively) must dynamically expand. This represents a formidable challenge for conventional deep learning approaches, which are typically designed for fixed input/output dimensions and suffer from catastrophic forgetting. Recognizing this inherent difficulty, we have turned to a paradigm shift in machine learning for inspiration, exploring approaches beyond traditional gradient-based optimization to fundamentally address these complexities.

\begin{figure*}[t]
\centering
% \vspace{-10pt}
\includegraphics[width=0.95\textwidth]{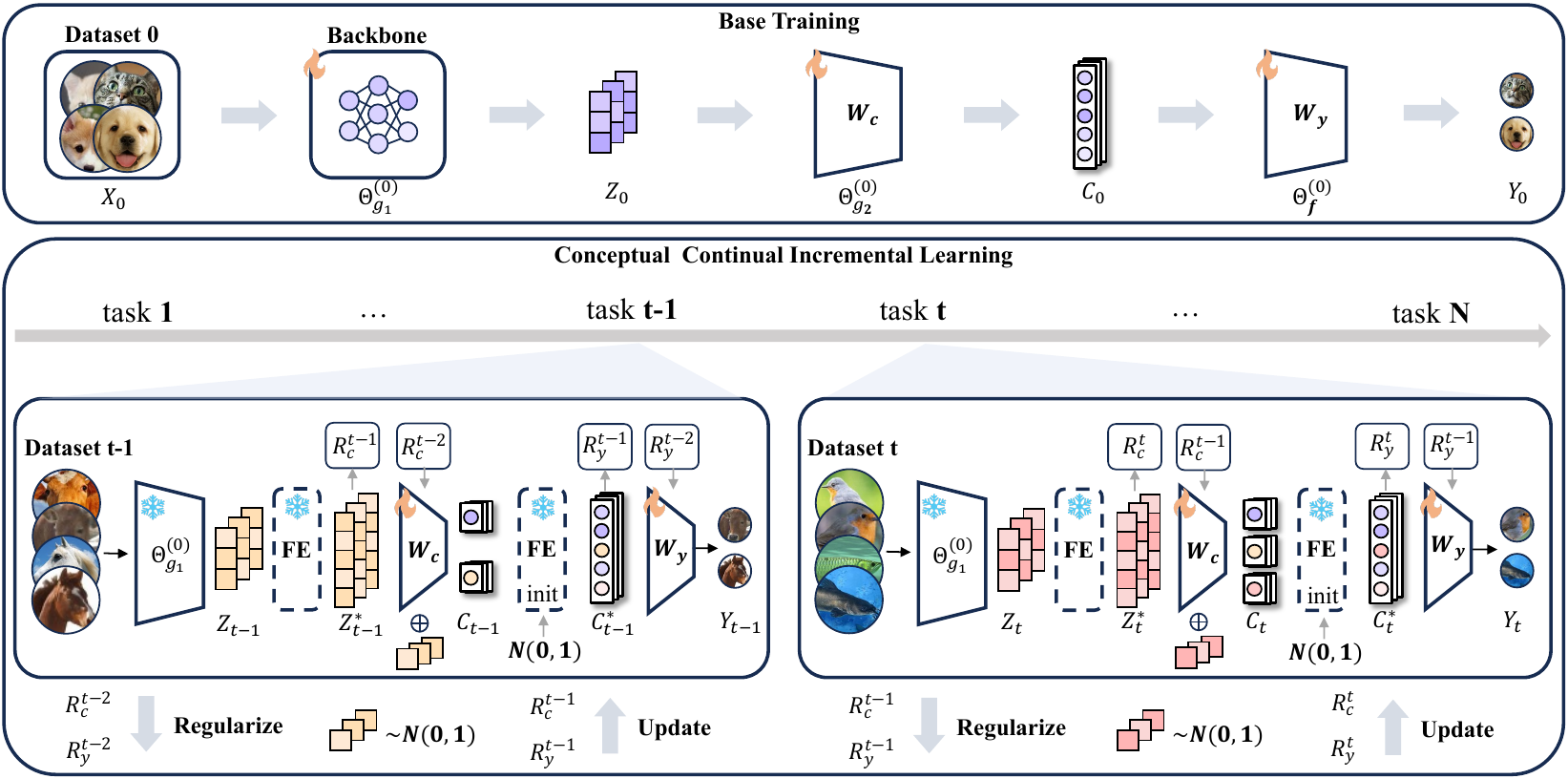}
% \vspace{-5pt}
\caption{Schematic overview of the \textbf{CONCIL framework} for continual learning in Multimodal Concept Bottleneck Models. The framework comprises two stages: an initial base training phase (Task 0, \textbf{Top Panel}) and subsequent incremental learning phases (Tasks $t \ge 1$, \textbf{Bottom Panel}). In base training, the backbone network, concept layer, and classifier are jointly optimized; the backbone is then frozen. For incremental tasks, CONCIL employs a novel recursive analytic approach: features ($Z_t$) are extracted, expanded ($Z_t^*$), and then used to recursively update both the concept layer weights ($W_c$) and classifier weights ($W_y$). Crucially, these updates dynamically accommodate the expansion of both concept and class dimensions (highlighted by orange blocks), integrating new knowledge without catastrophic forgetting. The recursive nature ensures absolute knowledge retention and computational efficiency by utilizing only current task data and previous summary matrices ($R_c^{t-1}, R_y^{t-1}$), thus avoiding the need to store or re-access prior raw data. This design is particularly beneficial for interpreting and adapting to dynamic multimodal data streams.}
\label{framework}
% \vspace{-15pt}
\end{figure*}

% \vspace{-7pt}
\section{Method}
\label{sec:method}
% \vspace{-3pt}

This section details our proposed continual learning framework, \textbf{CON}ceptual \textbf{C}ontinual \textbf{I}ncremental \textbf{L}earning (\textbf{CONCIL}). CONCIL is specifically designed for Concept Bottleneck Models (CBMs) to enable robust concept-incremental and class-incremental learning through a novel recursive analytic approach. This framework fundamentally addresses catastrophic forgetting by preserving historical knowledge without requiring the storage or re-access of prior data samples, making it highly suitable for dynamic data streams.

% \vspace{-7pt}
\subsection{Model Architecture and Notation}
% \vspace{-3pt}

Following our task definition, we extend the conventional CBM structure from $ \mathbf{x} \mapsto \mathbf{c} \mapsto y $ to $ \mathbf{x} \mapsto \mathbf{z} \mapsto \mathbf{c} \mapsto y $. This extension introduces an intermediate feature representation $ \mathbf{z} $ extracted by a backbone network, which is particularly relevant for handling complex raw inputs from various modalities. Formally:

%\begin{itemize}
\noindent \textbf{1)} $ \mathbf{x} \in \mathbb{R}^d $ denotes the input data. In multimodal contexts, $\mathbf{x}$ can represent a feature vector derived from diverse types of sources such as images, audio, or text, or even a concatenated representation of features from multiple modalities.

\noindent \textbf{2)} $ \mathbf{z} = g_1(\mathbf{x}; \Theta_{g_1}) \in \mathbb{R}^{d_z} $ is the high-level feature vector extracted by the backbone network $ g_1 $ with parameters $ \Theta_{g_1} $.

\noindent \textbf{3)} $ \mathbf{c} = g_2(\mathbf{z}; \Theta_{g_2}) \in \mathbb{R}^{L_{ \leq t}} $ represents the human-interpretable concept vector obtained via a concept mapping function $ g_2 $ with parameters $ \Theta_{g_2} $. Importantly, the dimension $L_{\leq t}$ of this concept space dynamically expands to accommodate new concepts introduced in task $t$. These concepts can represent cross-modal attributes derived from the input $\mathbf{x}$.

\noindent \textbf{4)} $ y = f(\mathbf{c}; \Theta_f) \in \mathbb{R}^{|Y_{ \leq t}|}$ is the final output predicted by a classifier $ f $ with parameters $ \Theta_f $. The output dimension $|Y_{ \leq t}|$ similarly grows to encompass all classes learned up to task $t$.
% \end{itemize}

At each learning phase $ t $ in a continual learning scenario, the model receives a new dataset $ \mathcal{D}_t = \{(\mathbf{x}_i, \mathbf{c}_i, y_i)\}_{i=1}^{N_t} $, where previously unseen concepts and classes emerge. Over multiple phases, this incremental addition of knowledge requires the model to not only learn new information but also rigorously retain what was learned in earlier phases, which is crucial for real-world multimodal understanding systems.

% \vspace{-7pt}
\subsection{Base Training and Feature Expansion}
% \vspace{-3pt}

In the initial training phase (base training, $t=0$), we jointly train the backbone $ g_1 $, concept layer $ g_2 $, and classifier $ f $ on the first dataset $ \mathcal{D}_0 $ using standard backpropagation (BP). This process yields initial parameter estimates $ \Theta_{g_1}^{(0)} $, $ \Theta_{g_2}^{(0)} $, and $ \Theta_f^{(0)} $. For an input $ \mathbf{x} $, the initial model output is:

% \vspace{-10pt}
\begin{equation}
    y = f(g_2(\mathbf{z}; \Theta_{g_2}^{(0)}); \Theta_f^{(0)}),
\end{equation}
% \vspace{-10pt}

\noindent where

% \vspace{-10pt}
\begin{equation}
\mathbf{z} = g_1(\mathbf{x}; \Theta_{g_1}^{(0)}).
\end{equation}
    % \vspace{-10pt}

After this initial phase, the backbone parameters $ \Theta_{g_1}^{(0)} $ are frozen as $ \Theta_{g_1}$. To enhance the feature space's capacity for analytic learning, we introduce a \textbf{Feature Expansion (FE)} transformation that maps the extracted feature $ \mathbf{z} $ to a higher-dimensional space $ \mathbf{z}^* $. This expanded feature representation $ \mathbf{z}^* $ facilitates robust concept separation in subsequent incremental phases by increasing the dimensionality of the representation space.

The expanded feature $ \mathbf{z}^* $ is defined as:

% \vspace{-10pt}
\begin{equation}
    \mathbf{z}^* = \phi(\mathbf{z}; W_{fe}) = \sigma(\mathbf{z} W_{fe}),
\end{equation}
% \vspace{-10pt}

\noindent where $ W_{fe} \in \mathbb{R}^{d_z \times d_{z^*}} $ is the expansion matrix, typically initialized with values from a normal distribution, and $ \sigma $ is a nonlinear activation function, such as ReLU. This feature expansion is applied consistently across all phases to provide a rich input space for the subsequent concept layer.

% \vspace{-7pt}
\subsection{Non-Recursive Solution for Concept-Incremental Learning}
% \vspace{-3pt}

With the expanded feature representation $ \mathbf{z}^* $, we formulate the concept mapping from $ \mathbf{z}^* $ to $\mathbf{c}$ as a linear regression problem, allowing for a closed-form analytic solution. For each phase $ t $, the objective is to learn the concept mapping matrix $ W_c^{(t)} \in \mathbb{R}^{d_{z^*} \times L_{ \leq t}}$. This matrix maps the expanded features to the cumulative concept space $L_{ \leq t}$. The objective minimizes the following regularized loss over the current task's data:

% \vspace{-15pt}
\begin{equation}
    \arg \min_{W_c^{(t)}} \| C_t - Z^{*}_t W_c^{(t)} \|_F^2 + \lambda_1 \| W_c^{(t)} \|_F^2,
\end{equation}
% \vspace{-5pt}

\noindent where $ C_t \in \mathbb{R}^{N_t \times L_{ \leq t}} $ is the matrix of concept labels for the $N_t$ samples in phase $t$. For concepts not present in the current task but part of $L_{ \leq t}$, their corresponding columns in $C_t$ are populated with zeros. $ Z^*_t \in \mathbb{R}^{N_t \times d_{{z}^*}} $ is the expanded feature matrix at phase $ t $, and $ \lambda_1 $ is a regularization parameter. This formulation yields the following closed-form solution for the full concept weight matrix at task $t$:

% \vspace{-15pt}
\begin{equation}
    \hat{W}_{c}^{(t)} = \left( \sum_{i=0}^{t} Z_i^{*T} Z_i^* + \lambda_1 I \right)^{-1} \left( \sum_{i=0}^{t} Z_i^{*T} C_i \right).
\end{equation}
% \vspace{-15pt}
%%%% c-->y

% \vspace{-3pt}
\subsection{Feature Expansion after Concept Mapping}
% \vspace{-3pt}

Following the concept mapping, we obtain the predicted concept representations $\hat{C_t} = Z^{*}_t W_c^{(t)} \in \mathbb{R}^{N_t \times L_{ \leq t}}$. To further enhance the discriminative power of the concept space for classification, we apply a second \textbf{Feature Expansion (FE)} transformation. This maps the extracted concept representation $ \hat{C_t} $ to a higher-dimensional space $ \hat{C_t^*} $. This expanded concept space representation $ \hat{C_t^*} $ is crucial for facilitating robust analytic learning in the subsequent classifier layer.

The expanded concept $ \hat{C_t}^* $ is defined as:

% \vspace{-10pt}
\begin{equation}
     \hat{C_t^*}  = \phi(\hat{C_t}; W_{fe}') = \sigma(\hat{C_t} W_{fe}'),
\end{equation}
if the updated concepts set is an empty set, the expanded concept would be:
\begin{equation}
    \hat{C}_t^* = \sigma\left( \begin{bmatrix} \hat{C}_t & \mathbf{0}_{N_t \times (L_{\max} - L_{\leq t})} \end{bmatrix} W_{fe}' \right),
\end{equation}
% \vspace{-10pt}

\noindent where $ W_{fe}' \in \mathbb{R}^{L_{\text{max}} \times d_{\hat{C_t}^*}} $ is the expansion matrix initialized with values from a normal distribution, and $ \sigma $ is a nonlinear activation function, such as ReLU. A crucial aspect here is that the input dimension for $W_{fe}'$ is the cumulative concept space $L_{ \leq t}$, which expands over phases. To handle this, $W_{fe}'$ is designed to accommodate the maximum expected concept dimension $L_{\text{max}}$, or it is dynamically expanded to match $L_{ \leq t}$ if $L_{ \leq t}$ exceeds previous dimensions, with zero-padding for current $\hat{C_t}$ if $L_{ \leq t}$ is less than $L_{\text{max}}$. This dynamic nature is critical for multimodal CBMs dealing with an ever-growing semantic concept vocabulary.

% \vspace{-5pt}
\subsection{Non-Recursive Solution for Class-Incremental Learning}
% \vspace{-3pt}

With the expanded concept representation $ \hat{C_t^*} $, we formulate the class prediction problem as a linear regression from $ \hat{C_t^*} $ to the target labels $ \mathbf{y} $, also using a closed-form analytic solution. For each phase $ t $, the objective is to learn the classifier matrix $ W_y^{(t)} \in \mathbb{R}^{d_{ \hat{C_t^*} } \times |Y_{\leq t}|}$, mapping the expanded concept features to the cumulative class label space $|Y_{\leq t}|$. This minimizes the following regularized loss:

% \vspace{-10pt}
\begin{equation}
    \arg \min_{W_y^{(t)}} \| Y_t - \hat{C_t^*} W_y^{(t)} \|_F^2 + \lambda_2 \| W_y^{(t)} \|_F^2,
\end{equation}
% \vspace{-5pt}

\noindent where $ Y_t \in \mathbb{R}^{N_t \times |Y_{\leq t}|} $ is the one-hot encoded matrix of class labels for the $N_t$ samples in phase $t$. Similar to $C_t$, columns corresponding to classes not present in task $t$ but part of $|Y_{\leq t}|$ are zeroed out. $\hat{C_t^*} \in \mathbb{R}^{N_t \times d_{\hat{C_t^*}}} $ is the expanded concept matrix at phase $ t $, and $ \lambda_2 $ is a regularization parameter. This formulation leads to the closed-form solution for the full classifier weight matrix:

% \vspace{-10pt}
\begin{equation}
    \hat{W}_{y}^{(t)} = \left( \sum_{i=0}^{t} \hat{C}_i^{*T} \hat{C}_i^* + \lambda_2 I \right)^{-1} \left( \sum_{i=0}^{t} \hat{C}_i^{*T} Y_i \right).
\end{equation}
% \vspace{-5pt}

% \vspace{-7pt}
\subsection{Conversion to Recursive Form}
% \vspace{-3pt}

To enable efficient learning in a continual setting, we convert these non-recursive closed-form solutions into a recursive update form. This recursive transformation is pivotal as it allows the model to incrementally update parameters using only the current phase’s data, thereby eliminating the need to store or re-access all prior data. This is a critical advantage for handling large-scale multimodal data streams under privacy and storage constraints.

Let $ \mathbf{R_c}^{(t)} = \left( \sum_{i=0}^{t} {Z}_i^{*T} {Z^*}_i + \lambda_1 I \right)^{-1} $ denote the regularized inverse correlation matrix for concept mapping at phase $ t $. Applying the Sherman-Morrison-Woodbury formula (also known as the matrix inversion lemma), we derive the recursive update for $ \mathbf{R_c}^{(t)} $. The detailed derivation can be found in Section \ref{sec:theoretical_derivation_of_recursive_update}. The recursive form for $ \mathbf{R_c}^{(t)} $ is:

% \vspace{-20pt}
\begin{equation}
    \resizebox{0.45\textwidth}{!}{$
\begin{aligned}
    &\mathbf{R_c}^{(t)} = \left( \sum_{i=0}^{t} {Z^*}_i^T Z^*_i + \lambda_1 I \right)^{-1} \\
    &= \mathbf{R_c}^{(t-1)} - \mathbf{R_c}^{(t-1)} {Z}_t^{*T} \left( I + {Z}^*_t \mathbf{R_c}^{(t-1)} {Z}_t^{*T} \right)^{-1} {Z}_t^* \mathbf{R_c}^{(t-1)}.
\end{aligned}
    $}
\end{equation}
% \vspace{-15pt}

Similarly, for $ \mathbf{R_y}^{(t)} = \left( \sum_{i=0}^{t} \hat{C_i^*}^{T} \hat{C_i^*} + \lambda_2 I \right)^{-1} $ which denotes the regularized inverse correlation matrix for class mapping at phase $ t $, the recursive form is:

% \vspace{-15pt}
\begin{equation}
    \resizebox{0.45\textwidth}{!}{$
\begin{aligned}
    &\mathbf{R_y}^{(t)} = \left( \sum_{i=0}^{t} \hat{C_i^*}^T \hat{C_i^*} + \lambda_2 I \right)^{-1} \\
    &= \mathbf{R_y}^{(t-1)} - \mathbf{R_y}^{(t-1)} \hat{C_t^*}^T \left( I + \hat{C_t^*} \mathbf{R_y}^{(t-1)} \hat{C_t^*}^T \right)^{-1} \hat{C_t^*} \mathbf{R_y}^{(t-1)}. 
\end{aligned}
    $}
\end{equation}
% \vspace{-15pt}

This derivation provides an efficient method for updating the inverse correlation matrices, allowing us to utilize the current phase data without recalculating the entire cumulative sum.

% \vspace{-7pt}
\subsection{Recursive Update for Weights}
% \vspace{-3pt}

Building upon the recursive forms for $ \mathbf{R_c}^{(t)} $ and $ \mathbf{R_y}^{(t)} $, we can similarly derive the recursive updates for the full concept layer weight matrix $ W_c^{(t)}$ and the full classifier weight matrix $ W_y^{(t)} $. This is where CONCIL explicitly handles the expansion of both concept and class dimensions. The updates are formulated as:

% \vspace{-13pt}
\begin{equation}
    \resizebox{0.45\textwidth}{!}{$
    W_c^{(t)} = \left[ W_c^{(t-1)} - \mathbf{R_c}^{(t)} {Z^*}_t^T ({Z^*}_t W_c^{(t-1)} - C_{t,\text{old}}) \quad \mathbf{R_c}^{(t)} {Z^*}_t^T C_{t,\text{new}} \right],
    $}
\end{equation}
% \vspace{-15pt}

\noindent where $C_{t,\text{old}}$ are the ground truth concepts for previously seen concepts in $C_t$ and $C_{t,\text{new}}$ are the ground truth concepts for new concepts in $C_t$. The term $W_c^{(t-1)} - \mathbf{R_c}^{(t)} {Z^*}_t^T ({Z^*}_t W_c^{(t-1)} - C_{t,\text{old}})$ updates the weights corresponding to previously learned concepts, while $\mathbf{R_c}^{(t)} {Z^*}_t^T C_{t,\text{new}}$ computes the weights for the newly introduced concepts. The final $W_c^{(t)}$ is a column concatenation of these two parts, effectively expanding the concept layer's output dimension.

Similarly, the classifier weight update can be formulated as:

% \vspace{-13pt}
\begin{equation}
    \resizebox{0.45\textwidth}{!}{$
    W_y^{(t)} = \left[ W_y^{(t-1)} - \mathbf{R_y}^{(t)} \hat{C_t^*}^T (\hat{C_t^*} W_y^{(t-1)} - Y_{t,\text{old}}) \quad \mathbf{R_y}^{(t)} \hat{C_t^*}^T Y_{t,\text{new}} \right],
    $}
\end{equation}
% \vspace{-15pt}

\noindent where $Y_{t,\text{old}}$ and $Y_{t,\text{new}}$ are the ground truth labels for old and new classes within $Y_t$, respectively. This allows for both updating existing class prediction capabilities and incorporating new classes through column concatenation. This unique recursive formulation directly addresses the growing output dimensions inherent in concept- and class-incremental learning, a significant challenge for conventional methods, particularly in multimodal classification tasks.

% \vspace{-8pt}
\subsection{Properties of the Recursive Framework}
% \vspace{-3pt}

The CONCIL framework, built upon these recursive analytic updates, offers several compelling advantages:

\noindent \textbf{(i) Absolute Knowledge Memory:} By maintaining the recursive update form derived from the closed-form solution of linear regression, CONCIL theoretically guarantees that the model fully retains knowledge from all previous phases. This means the model behaves as though it had been trained on a centralized, comprehensive dataset encompassing all past and current data, effectively achieving "absolute knowledge memory" and eliminating catastrophic forgetting.

\noindent \textbf{(ii) Privacy Protection:} Since the recursive update relies only on the current phase data and the summary inverse correlation matrices $ \mathbf{R_c}$ and $ \mathbf{R_y} $ (which do not contain sensitive raw data), it does not require storage of previous phase raw data. This inherent design protects data privacy, a crucial consideration for real-world deployments involving sensitive multimodal data.

\noindent \textbf{(iii) Computational Efficiency and Scalability:} The recursive framework significantly reduces the computational load by avoiding redundant recalculations on historical data. Each update involves only matrix multiplications and inversions of relatively small matrices (dependent on feature/concept dimensions, not total data size), making it highly suitable for real-time continual learning scenarios and large-scale multimodal data processing.

\noindent \textbf{(iv) Multimodal Adaptability:} The CBM architecture naturally facilitates multimodal learning by mapping diverse raw inputs to a unified, interpretable concept space. CONCIL's ability to incrementally expand this concept space and the subsequent class space, while preserving knowledge, makes it uniquely adaptable to evolving multimodal feature sets and new cross-modal concepts that frequently emerge in dynamic real-world environments.
%% \vspace{-5pt}

\begin{table*}[t]
% \vspace{-15pt}
    \centering
    \small % 调整字体大小
    \setlength\tabcolsep{10pt} % 减小列间距
    \caption{Comparison of Model Performance Across Incremental Phases: Concept and Class Accuracy, and Average Forgetting Rates for CUB and AwA Datasets. Higher accuracy and lower forgetting rates indicate superior performance. The underlined values represent the best performance for each metric and phase.}
    % \vspace{-10pt}
    \label{CONCIL_table}
        \begin{adjustbox}{max width=0.95\textwidth} % 调整表格宽度
    \begin{tabular}{lccccccccc}
        \toprule
        \textbf{Metric} & \textbf{Phase 2} & \textbf{Phase 3} & \textbf{Phase 4} & \textbf{Phase 5} & \textbf{Phase 6} & \textbf{Phase 7} & \textbf{Phase 8} & \textbf{Phase 9} & \textbf{Average} \\
        \midrule
        \multicolumn{10}{c}{\textbf{Average Concept Accuracy $\uparrow$ (CUB)}} \\
        \cmidrule(r){1-10}
        \textbf{Baseline} & 0.7357 & 0.7286 & 0.7046 & 0.6930 & 0.6760 & 0.6638 & 0.6647 & 0.6502 & 0.6896 \\
        \rowcolor{gray!20} % 设置CONCIL行的背景颜色
        \textbf{CONCIL} & \underline{0.8233} & \underline{0.8220} & \underline{0.8200} & \underline{0.8207} & \underline{0.8205} & \underline{0.8202} & \underline{0.8203} & \underline{0.8204} & \underline{0.8209} \\
        \midrule
        \multicolumn{10}{c}{\textbf{Average Class Accuracy $\uparrow$ (CUB)}} \\
        \cmidrule(r){1-10}
        \textbf{Baseline} & 0.6119 & 0.4297 & 0.3305 & 0.2692 & 0.2263 & 0.1950 & 0.1723 & 0.1513 & 0.2983 \\
        \rowcolor{gray!20} % 设置CONCIL行的背景颜色
        \textbf{CONCIL} & \underline{0.6287} & \underline{0.6216} & \underline{0.6163} & \underline{0.6064} & \underline{0.6090} & \underline{0.6118} & \underline{0.6079} & \underline{0.6043} & \underline{0.6133} \\
        \midrule
        \multicolumn{10}{c}{\textbf{Average Concept Accuracy $\uparrow$ (AwA)}} \\
        \cmidrule(r){1-10}
        \textbf{Baseline} & 0.9262 & 0.8709 & 0.8364 & 0.8095 & 0.7866 & 0.7747 & 0.7592 & 0.7488 & 0.8140 \\
        \rowcolor{gray!20} % 设置CONCIL行的背景颜色
        \textbf{CONCIL} & \underline{0.9708} & \underline{0.9699} & \underline{0.9704} & \underline{0.9701} & \underline{0.9699} & \underline{0.9703} & \underline{0.9704} & \underline{0.9702} & \underline{0.9703} \\
        \midrule
        \multicolumn{10}{c}{\textbf{Average Class Accuracy $\uparrow$ (AwA)}} \\
        \cmidrule(r){1-10}
        \textbf{Baseline} & 0.7601 & 0.5036 & 0.3888 & 0.3149 & 0.2644 & 0.2283 & 0.2010 & 0.1794 & 0.3550 \\
        \rowcolor{gray!20} % 设置CONCIL行的背景颜色
        \textbf{CONCIL} & \underline{0.8739} & \underline{0.8675} & \underline{0.8647} & \underline{0.8624} & \underline{0.8580} & \underline{0.8616} & \underline{0.8561} & \underline{0.8550} & \underline{0.8624} \\
        \midrule
        \multicolumn{10}{c}{\textbf{Average Concept Forget Rate $\downarrow$ (CUB)}} \\
        \cmidrule(r){1-10}
        \textbf{Baseline} & \underline{-0.0490} & \underline{-0.0347} & \underline{-0.0109} & \underline{-0.0032} & 0.0159 & 0.0032 & 0.0006 & 0.0122 & \underline{-0.0082} \\
        \rowcolor{gray!20} % 设置CONCIL行的背景颜色
        \textbf{CONCIL} & -0.0008 & -0.0006 & -0.0008 & -0.0003 &\underline{ 0.0003} & \underline{-0.0006} & \underline{-0.0003} & \underline{-0.0004} & -0.0004 \\
        \midrule
        \multicolumn{10}{c}{\textbf{Average Class Forget Rate $\downarrow$ (CUB)}} \\
        \cmidrule(r){1-10}
        \textbf{Baseline} & 0.8101 & 0.7922 & 0.8000 & 0.8168 & 0.8254 & 0.8279 & 0.8407 & 0.8093 & 0.8153 \\
        \rowcolor{gray!20} % 设置CONCIL行的背景颜色
        \textbf{CONCIL} & \underline{0.1239} & \underline{0.0610} & \underline{0.0847} & \underline{0.0948} & \underline{0.0843} & \underline{0.0938} & \underline{0.0945} & \underline{0.0978} & \underline{0.0919} \\
        \midrule
        \multicolumn{10}{c}{\textbf{Average Concept Forget Rate $\downarrow$ (AwA)}} \\
        \cmidrule(r){1-10}
        \textbf{Baseline} & 0.2158 & 0.3588 & 0.2530 & 0.2502 & 0.2554 & 0.2635 & 0.3470 & 0.2685 & 0.2765 \\
        \rowcolor{gray!20} % 设置CONCIL行的背景颜色
        \textbf{CONCIL} & \underline{0.0073} & \underline{0.0048} & \underline{0.0043} & \underline{0.0036} & \underline{0.0034} & \underline{0.0032} & \underline{0.0038} & \underline{0.0033} & \underline{0.0042} \\
        \midrule
        \multicolumn{10}{c}{\textbf{Average Class Forget Rate $\downarrow$ (AwA)}} \\
        \cmidrule(r){1-10}
        \textbf{Baseline} & 0.9239 & 0.9273 & 0.9503 & 0.9519 & 0.9559 & 0.9619 & 0.9607 & 0.8603 & 0.9365 \\
        \rowcolor{gray!20} % 设置CONCIL行的背景颜色
        \textbf{CONCIL} & \underline{0.1704} & \underline{0.1045} & \underline{0.0943} & \underline{0.0808} & \underline{0.0791} & \underline{0.0692} & \underline{0.1449} & \underline{0.0803} & \underline{0.1029} \\
        \bottomrule
    \end{tabular}
      \end{adjustbox}
      % \vspace{-10pt}
\end{table*}

% \vspace{-7pt}
\section{Experiments and Results}
% \vspace{-4pt}

\subsection{Datasets and Backbone}
% \vspace{-3pt}
This section outlines the datasets and the backbone architecture employed in our experiments. Our study focuses on evaluating the effectiveness of our proposed method on two well-known benchmark datasets: the Caltech-UCSD Birds-200-2011 (CUB) \cite{wah2011caltech} and the Animals with Attributes (AwA) \cite{xian2018zero} datasets. Both datasets are widely used in visual recognition and concept learning, providing a rich testbed for multimodal concept-based models by linking image data to human-interpretable attributes.

\noindent \textbf{CUB Dataset.} The Caltech-UCSD Birds-200-2011 (CUB) dataset \cite{wah2011caltech} is specifically tailored for the task of fine-grained bird classification, featuring 11,788 images representing 200 distinct species. Accompanying these images are 312 binary attributes, which provide rich, high-level semantic descriptions of the birds (e.g., 'has\_belly\_color\_white', 'shape\_of\_beak\_curved'). These attributes are crucial for enabling concept-based interpretability and represent a cross-modal link between visual input and symbolic understanding.

\noindent \textbf{AwA Dataset.} The AwA dataset consists of 37,322 images from 50 animal categories, each annotated with 85 binary attributes (e.g., 'has\_fur', 'walks', 'quadruped'). This dataset similarly offers a strong foundation for evaluating multimodal CBMs, as it connects visual animal features to semantic attributes. We divided the dataset into training and testing sets, with an equal distribution of images per class, resulting in 18,652 training images and 18,670 testing images. No modifications were made to the binary attributes, preserving the original annotations for both training and evaluation.

\noindent \textbf{Backbone Architecture.} To serve as the feature extractor for our experiments, we utilized a pre-trained ResNet50 model as the backbone. This widely adopted architecture has proven effective in learning robust visual features from diverse image data.

% \vspace{-13pt}
\subsection{Setting}
% \vspace{-3pt}
%%% 描述一下任务设定
%% 对CUB数据集和AwA数据集进行处理，使其符合概念增量和类别增量的持续学习的任务设定。首先在phase1，可以获得前n%的类别数据和这前n%类别数据里的前m%的概念的访问权限。将其设置为p个phase的持续学习，则从第二个phase开始，每一个phase会获得（1-n%)/（p-1）的新类别数据，且这些新类别数据会可以访问多（1-m%)/（p-1)的概念。也就是说，后续的phase，会见到新的类别数据，且可以访问的concept权限越来越多。同时对于每个phase，只能访问当前phase的数据和上一phase的参数权重。

To rigorously evaluate the performance of our proposed Concept-Incremental and Class-Incremental Continual Learning (CICIL) framework for CBMs, we designed a series of experiments on the CUB and AwA datasets. These datasets were carefully adapted to simulate the concept-incremental and class-incremental continual learning paradigm, reflecting real-world scenarios where *multimodal data streams evolve over time. The experimental setup is structured into multiple sequential phases, each corresponding to a distinct learning task with a progressively increasing number of classes and associated concepts.

\noindent \textbf{Phase-wise Data Splitting and Access Control.} For both the CUB and AwA datasets, the initial phase (Phase 1) is configured to provide access to only the first $n\%$ of the total classes and the first $m\%$ of the concepts directly associated with these classes. Specifically, in Phase 1, the model is trained on a subset of the data that includes the earliest $n\%$ of the classes and their corresponding $m\%$ of concepts. This initial setup allows the model to establish a foundational understanding of the problem domain and its initial concept-attribute relationships before being exposed to new information.

Subsequent phases are designed to incrementally introduce new classes and expand the concept vocabulary. From Phase 2 onwards, each phase incorporates an additional $ \frac{1-n\%}{p-1} $ of the remaining classes and an additional $ \frac{1-m\%}{p-1} $ of the concepts, where $ p $ denotes the total number of phases. This gradual increase ensures that the model is continuously challenged to learn novel information while strictly maintaining the stability of previously acquired knowledge. It is crucial to note that for each phase, the model has access only to the current phase's data and the parameters from the immediately preceding phase. This constraint rigorously simulates realistic deployment scenarios where storing and revisiting all past raw data might be infeasible due to practical considerations such as privacy regulations and storage limitations.

In our experiments, we set $n=50$, $m=50$, and $p$ to range from 2 to 9, allowing for a comprehensive evaluation across multiple incremental tasks.

\begin{figure*}[t!]
\centering
% \vspace{-10pt}
\includegraphics[width=0.9\textwidth]{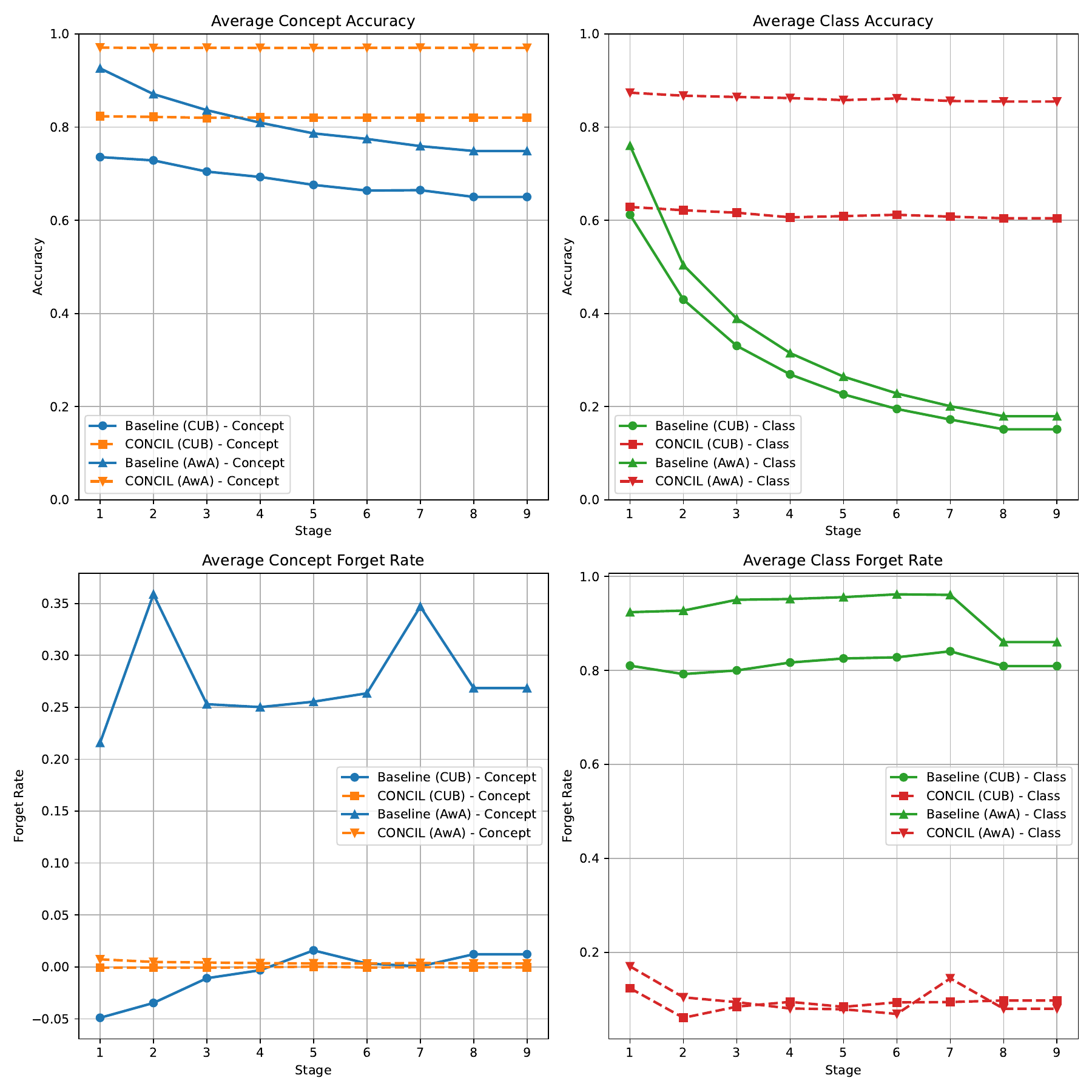}
% \vspace{-15pt}
\caption{Performance Comparison of Baseline and CONCIL Models on CUB and AwA Datasets across Incremental Phases. The top row presents the average concept accuracy (left) and average class accuracy (right) for both Baseline and CONCIL models. The bottom row displays the average concept forgetting rate (left) and average class forgetting rate (right). Each plot visualizes metrics across different phases, compellingly highlighting CONCIL's superior performance, significantly higher accuracy, and remarkably lower forgetting rates compared to the Baseline model in a continual learning setting, particularly relevant for dynamic multimodal data understanding.}
\label{visual-CONCIL_ALL}
% \vspace{-20pt}
\end{figure*}

% \vspace{-5pt}
\subsection{Metrics}

To comprehensively evaluate the performance of our CICIL framework, we employ four key metrics that collectively measure both the accuracy and the stability of the model across different phases. These metrics provide a balanced assessment of the model's ability to learn new information while retaining previously acquired knowledge, particularly critical for understanding the behavior of multimodal concept-based models in dynamic environments. The \textbf{Average Concept Accuracy} ($ A_{\text{concept}}(t) $) and \textbf{Average Class Accuracy} ($ A_{\text{class}}(t) $) quantify the mean accuracy of concept and class predictions, respectively, across all tasks learned up to the current phase $ t $:

% \vspace{-15pt}
\begin{equation}
    A_{\text{concept}}(t) = \frac{1}{t} \sum_{k=1}^{t} \text{Accuracy}(\mathcal{D}_k^{\text{test}}, \mathbf{c}),
\end{equation}
% \vspace{-10pt}
\begin{equation}
    A_{\text{class}}(t) = \frac{1}{t} \sum_{k=1}^{t} \text{Accuracy}(\mathcal{D}_k^{\text{test}}, y).
\end{equation}
% \vspace{-10pt}

%% \vspace{-5pt}

The \textbf{Average Concept Forgetting Rate} ($ F_{\text{concept}}(t) $) and \textbf{Average Class Forgetting Rate} ($ F_{\text{class}}(t) $) measure the mean rate at which the model forgets previously learned concepts and classes, respectively, across all tasks up to the current phase $ t $:
% \vspace{-5pt}
\begin{equation}
\resizebox{0.5\textwidth}{!}{$
    F_{\text{concept}}(t) = \frac{1}{t-1} \sum_{k=1}^{t-1} \left( \max_{1 \leq j < t} \text{Accuracy}(\mathcal{D}_k^{\text{test}}, \mathbf{c}) - \text{Accuracy}(\mathcal{D}_k^{\text{test}}, \mathbf{c})_t \right),
    $}
\end{equation}

% \vspace{-10pt}
\begin{equation}
\resizebox{0.5\textwidth}{!}{$
   F_{\text{class}}(t) = \frac{1}{t-1} \sum_{k=1}^{t-1} \left( \max_{1 \leq j < t} \text{Accuracy}(\mathcal{D}_k^{\text{test}}, y) - \text{Accuracy}(\mathcal{D}_k^{\text{test}}, y)_t \right).
    $}
\end{equation}
%% \vspace{-5pt}

These metrics provide a balanced evaluation of the model's ability to learn new information while retaining previously acquired knowledge, which is essential for robust continual learning.

% \vspace{-5pt}
\subsection{Setup}
% \vspace{-3pt}
We configured the experimental setup with the following parameters and techniques. We set $\lambda_1 = 500$ , $\lambda_2 = 1$, $d_{{z}^*} = 25000$, and $d_{\hat{C_t}^*} = 25000$. The initial learning rate for the backbone was set to $1 \times 10^{-4}$, and a weight decay of $5 \times 10^{-5}$ was applied to prevent overfitting during its base training. An exponential learning rate scheduler with a decay factor ($\gamma$) of 0.95 was used to enhance the backbone's generalization ability. The concept loss, crucial for aligning learned features with human-interpretable concepts, was weighted by a factor of 0.5 to balance the primary task loss during base training. During the backbone's base training phase and for data preparation in each phase of the CICIL task framework, the datasets were augmented with random color jittering, random horizontal flipping, and random cropping to a resolution of 256 pixels, following the guidelines established by Koh et al. \cite{koh2020concept} with minor adjustments to the resolution. For inference, images were center-cropped and resized to 256 pixels to ensure consistency with the training input size. Each phase of the CICIL task framework (i.e., the analytic updates for $W_c$ and $W_y$) was performed in a single epoch, reflecting the nature of the task as a linear fitting problem, while the baseline model (representing a conventional CBM re-trained incrementally without explicit forgetting mitigation) was trained for 50 epochs per phase, consistent with typical CBM training setups. All experiments were conducted on an A800 GPU to provide the necessary computational power for efficient processing.

% \vspace{-5pt}
\subsection{Experimental Results and Analysis}
% \vspace{-3pt}

The results presented in Table~\ref{CONCIL_table} compellingly demonstrate the significant superiority of our proposed method, CONCIL, over the baseline model on both the CUB and AwA datasets. This comprehensive evaluation highlights CONCIL's efficacy in handling the complex multimodal concept-incremental and class-incremental learning task.

Specifically, for average concept accuracy, CONCIL consistently outperforms the baseline. It achieves an impressive 19\% improvement on the CUB dataset (from 0.6896 to 0.8209) and a substantial 19.2\% improvement on the AwA dataset (from 0.8140 to 0.9703). This consistent and considerable gain across two distinct fine-grained image datasets underscores CONCIL's robustness and generalizability in accurately identifying and retaining a growing vocabulary of human-interpretable concepts.

For average class accuracy, the performance gap is even more pronounced, directly showcasing CONCIL's remarkable ability to mitigate catastrophic forgetting for classification tasks. On the CUB dataset, CONCIL achieves an improvement of approximately 31.5 percentage points (from 0.2983 to 0.6133) over the baseline. On the AwA dataset, this improvement is even more significant at approximately 50.7 percentage points (from 0.3550 to 0.8624). These substantial gains in class accuracy decisively highlight the effectiveness of CONCIL in handling complex incremental classification tasks, particularly where the classifier's output dimension also continually expands to accommodate new classes.

Moreover, CONCIL drastically reduces the forgetting rate, a critical metric in continual learning that directly reflects knowledge retention. The average class forgetting rate decreases by an astounding 88.7\% on the CUB dataset (from 0.8153 to 0.0919) and by 89\% on the AwA dataset (from 0.9365 to 0.1029). This remarkable reduction in forgetting rate indicates that CONCIL not only excels in learning new concepts and classes but also effectively retains previously learned information, thereby fundamentally addressing one of the primary and most challenging problems in continual learning. These results empirically validate our theoretical claim of "absolute knowledge memory" for the analytic layers.

To gain a more intuitive understanding of these profound performance differences, Figure~\ref{visual-CONCIL_ALL} provides a visual comparison of the baseline and CONCIL models' performance trajectory across phases. As the number of training steps (phases) increases, the average accuracy of the baseline model drops significantly, while its average forgetting rate gradually increases. This trend is particularly evident in the later phases of training, where the baseline model's performance deteriorates markedly, typical of catastrophic forgetting in conventional deep learning setups.

In stark contrast, the CONCIL model maintains a remarkably stable and high level of performance throughout the entire continual learning process. It consistently achieves substantially higher average accuracy and exhibits a significantly lower forgetting rate compared to the baseline. This unparalleled stability and consistency in performance across different phases highlight the superior capabilities of CONCIL in effectively managing the inherent trade-off between learning new tasks and robustly retaining old knowledge, an essential characteristic for robust multimodal AI systems deployed in dynamic environments.

Furthermore, the detailed visualizations in Figure~\ref{visual-CONCIL_1} in Appendix offer additional insights into the performance dynamics of both models across individual concepts and classes per phase. It is patently clear that CONCIL consistently outperforms the baseline in both concept and class accuracy, maintaining exceptionally high levels of performance even as the complexity and diversity of the multimodal concept-attribute tasks increase. This visual evidence strongly supports the quantitative findings in Table~\ref{CONCIL_table} and reinforces CONCIL's effectiveness.

\section{Limitations and Future Work}
\label{sec:limitation}

While the proposed CONceptual Continual Incremental Learning (CONCIL) framework demonstrates significant advancements in addressing the challenges of continual learning for Multimodal Concept Bottleneck Models (CBMs), certain limitations warrant consideration and present compelling avenues for future research.

Firstly, while CONCIL effectively prevents catastrophic forgetting through its innovative use of recursive matrix operations, its inherent reliance on linear regression for updating concept and decision layers might constrain its capacity to capture highly complex, intricate non-linear relationships that could emerge between concepts and classes. This limitation could become more pronounced as the overall number or the semantic complexity of concepts and classes grows, particularly in highly granular multimodal understanding tasks. Future work will explore strategies to introduce non-linearity while preserving the benefits of analytic updates.

Secondly, the current framework assumes a structured, incremental introduction of new concepts and classes across phases. However, real-world multimodal data streams are often characterized by abrupt shifts in data distribution and the unpredictable emergence of novel information or complex co-occurrence patterns across modalities. Adapting CONCIL to robustly handle such highly dynamic, unstructured, and noisy changes in concept and class availability remains an important open challenge for enhanced real-world applicability.

Lastly, while CONCIL is designed to be computationally efficient compared to conventional gradient-based continual learning methods, the overhead associated with maintaining and updating the inverse correlation matrices for an extremely vast number of concepts or classes might become substantial. This could potentially limit its scalability in truly hyperscale multimodal applications demanding an exceptionally high dimensionality of concepts or classes. Optimizing memory and computational footprint for such extreme scenarios will be a focus of future research.

Furthermore, while CBMs are fundamentally designed for human interpretability, the introduction of intermediate feature expansion layers ($Z^*$ and $\hat{C}^*$) in CONCIL, while crucial for analytic learning, might introduce an additional layer of abstraction. Although the final concepts remain interpretable, the direct traceability of how raw multimodal input maps to specific concepts through these expanded representations could potentially become less intuitive. This opens avenues for future research to develop new algorithmic approaches that ensure maximal transparency and direct conceptual traceability throughout the entire continual learning process, thereby ensuring that the interpretability aspect of CBMs remains fully uncompromised, even in highly dynamic and expanding multimodal concept spaces.
% \vspace{-5pt}
\section{Conclusion}
% \vspace{-3pt}

\textbf{CON}ceptual \textbf{C}ontinual \textbf{I}ncremental \textbf{L}earning (\textbf{CONCIL}) represents a pioneering recursive analytic framework for Multimodal Concept Bottleneck Models, effectively eliminating catastrophic forgetting. By reformulating concept and decision layer updates as linear regression problems, CONCIL achieves “absolute knowledge memory” and significant computational efficiency. Experimental results compellingly validate its superior performance in both concept- and class-incremental settings, establishing a new paradigm for adaptive and interpretable AI in dynamic multimodal data streams. Future work will focus on extending its capabilities to capture more complex non-linear relationships, handle abrupt data shifts, and enhance scalability for even larger-scale multimodal AI applications.

\section{Appendix}

\subsection{Theoretical Derivation of Recursive Update}
\label{sec:theoretical_derivation_of_recursive_update}

In this section, we detail the theoretical derivation of the recursive update for the regularized inverse correlation matrix $ \mathbf{R_c}^{(t)} $ and $ \mathbf{R_y}^{(t)} $. This derivation forms the mathematical backbone of CONCIL's ability to achieve absolute knowledge retention.

The Sherman-Morrison-Woodbury formula (or matrix inversion lemma) provides the foundation for transforming the non-recursive solution into a recursive one. It states that for any invertible matrix $ A $, and matrices $ U, C, V $ of compatible dimensions, where $ C $ is invertible, we have:

\begin{equation}
\resizebox{0.45\textwidth}{!}{$
    (A + UCV)^{-1} = A^{-1} - A^{-1} U (C^{-1} + V A^{-1} U)^{-1} V A^{-1}.
    $}
\end{equation}

Applying this to our concept mapping problem, we let:

\begin{equation}
  A = \sum_{i=0}^{t-1} {Z^*}_i^T {Z^*}_i + \lambda_1 I = (\mathbf{R_c}^{(t-1)})^{-1},
\end{equation}

\noindent and we want to find $ \mathbf{R_c}^{(t)} = \left( \sum_{i=0}^{t} {Z^*}_i^T {Z^*}_i + \lambda_1 I \right)^{-1} = (A + {Z^*}_t^T Z^*_t)^{-1}$.

Here, we set $ U = {Z^*}_t^T $, $ C = I $ (an identity matrix of appropriate size), and $ V = {Z^*}_t $. Applying the lemma:

\begin{equation}
   \mathbf{R_c}^{(t)} = A^{-1} - A^{-1} {Z^*}_t^T (I^{-1} + {Z^*}_t A^{-1} {Z^*}_t^T)^{-1} {Z^*}_t A^{-1}. 
\end{equation}

Recognizing $ A^{-1} $ as $ \mathbf{R_c}^{(t-1)} $, we derive the recursive form:

\begin{equation}
\resizebox{0.45\textwidth}{!}{$
    \mathbf{R_c}^{(t)} = \mathbf{R_c}^{(t-1)} - \mathbf{R_c}^{(t-1)} {Z^*}_t^T (I + {Z^*}_t \mathbf{R_c}^{(t-1)} {Z^*}_t^T)^{-1} {Z^*}_t \mathbf{R_c}^{(t-1)}.
    $}
\end{equation}

A similar derivation applies for $ \mathbf{R_y}^{(t)} $ using $\hat{C_t^*}$ instead of $Z_t^*$. This theoretical foundation proves how to update the correlation matrices recursively based solely on the current phase data, thus maintaining computational efficiency while ensuring absolute knowledge retention.

\begin{figure*}[t!]
\centering
%% \vspace{-10pt}
\includegraphics[width=0.95\textwidth]{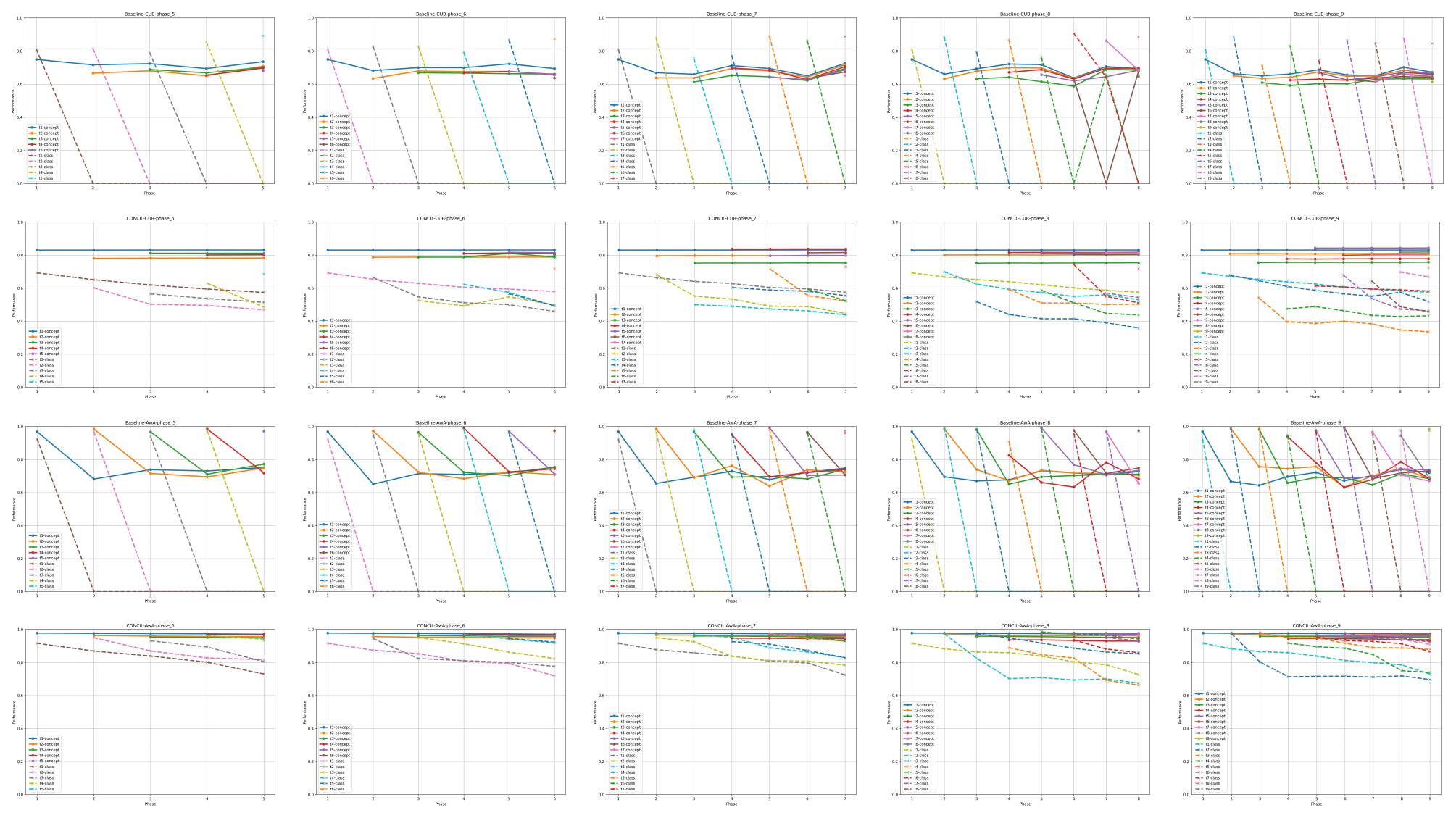}
%% \vspace{-15pt}
\caption{Visualization of Concept and Class Accuracy Across Phases for Baseline and CONCIL. The first and third rows illustrate the performance of the Baseline model on the CUB and AwA datasets, respectively. The second and fourth rows display the corresponding results for the proposed CONCIL model. Each column represents a sequential phase, demonstrating how CONCIL maintains high performance and resists catastrophic forgetting in evolving multimodal concept-attribute scenarios.}
\label{visual-CONCIL_1}
%% \vspace{-10pt}
\end{figure*}

\section{Limitations and Future Work}
\label{sec:limitation}

While the proposed CONceptual Continual Incremental Learning (CONCIL) framework demonstrates significant advancements in addressing the challenges of continual learning for Multimodal Concept Bottleneck Models (CBMs), certain limitations warrant consideration and present compelling avenues for future research.

Firstly, while CONCIL effectively prevents catastrophic forgetting through its innovative use of recursive matrix operations, its inherent reliance on linear regression for updating concept and decision layers might constrain its capacity to capture highly complex, intricate non-linear relationships that could emerge between concepts and classes. This limitation could become more pronounced as the overall number or the semantic complexity of concepts and classes grows, particularly in highly granular multimodal understanding tasks. Future work will explore strategies to introduce non-linearity while preserving the benefits of analytic updates.

Secondly, the current framework assumes a structured, incremental introduction of new concepts and classes across phases. However, real-world multimodal data streams are often characterized by abrupt shifts in data distribution and the unpredictable emergence of novel information or complex co-occurrence patterns across modalities. Adapting CONCIL to robustly handle such highly dynamic, unstructured, and noisy changes in concept and class availability remains an important open challenge for enhanced real-world applicability.

Lastly, while CONCIL is designed to be computationally efficient compared to conventional gradient-based continual learning methods, the overhead associated with maintaining and updating the inverse correlation matrices for an extremely vast number of concepts or classes might become substantial. This could potentially limit its scalability in truly hyperscale multimodal applications demanding an exceptionally high dimensionality of concepts or classes. Optimizing memory and computational footprint for such extreme scenarios will be a focus of future research.

Furthermore, while CBMs are fundamentally designed for human interpretability, the introduction of intermediate feature expansion layers ($Z^*$ and $\hat{C}^*$) in CONCIL, while crucial for analytic learning, might introduce an additional layer of abstraction. Although the final concepts remain interpretable, the direct traceability of how raw multimodal input maps to specific concepts through these expanded representations could potentially become less intuitive. This opens avenues for future research to develop new algorithmic approaches that ensure maximal transparency and direct conceptual traceability throughout the entire continual learning process, thereby ensuring that the interpretability aspect of CBMs remains fully uncompromised, even in highly dynamic and expanding multimodal concept spaces.

%Bibliography
\bibliographystyle{unsrt}  
\bibliography{references}  

\end{document}